\documentclass{article}

\usepackage[preprint]{neurips_2020}
\usepackage{microtype}
\usepackage[pdftex]{graphicx}
\usepackage{booktabs} 

\usepackage[utf8]{inputenc} 
\usepackage[T1]{fontenc}    
\usepackage{url}            
\usepackage{amsfonts}       
\usepackage{nicefrac}       
\usepackage[inline]{enumitem}

\usepackage{wrapfig, blindtext}
\usepackage{epstopdf}
\usepackage{booktabs} 
\usepackage{extarrows}

\usepackage{amsmath}   
\usepackage{amssymb}
\usepackage{wasysym}
\usepackage{multirow}
\usepackage{amsthm}
\usepackage{bbm}
\usepackage{tikz}
\usepackage{enumerate}
\usepackage{footmisc}
\usepackage{algorithm}
\usepackage{algorithmic}
\usepackage{setspace}
\usepackage{caption}
\usepackage{makecell}

\usepackage{natbib}
\usepackage[utf8]{inputenc} 
\usepackage[T1]{fontenc}    
\usepackage{hyperref}       
\usepackage{url}            
\usepackage{booktabs}       
\usepackage{graphicx}
\usepackage{wrapfig,lipsum,booktabs}
\usepackage{subcaption}
\usepackage{amsmath}
\usepackage{tikz}
\usepackage{graphicx}
\usetikzlibrary{positioning}

\usepackage{amsfonts}
\usepackage{mathtools}
\usepackage{framed}
\usepackage{microtype}

\usepackage{siunitx}
\usepackage{xr-hyper}
\usepackage{xr}
\usepackage{xr}
\usepackage{wrapfig}
\usepackage{amsmath,amsthm}
\usepackage{amssymb}
\usepackage{epstopdf}
\usepackage{mathtools}

\usepackage{hyperref}

\definecolor{mygray}{gray}{0.5}
\renewcommand{\algorithmiccomment}[1]{\;\;\;\textcolor{mygray}{$//$\;\textit{#1}}}
\makeatletter

\newcommand{\mm}{\textit{Manifold Mixup }}
\newcommand{\graphmix}{GraphMix }
\newcommand{\graphmixns}{GraphMix}
\newcommand{\graphmixgcn}{GraphMix(GCN) }
\newcommand{\graphmixgat}{GraphMix(GAT) }

\makeatother

\DeclareMathOperator{\betadist}{Beta}

\DeclareMathOperator{\sharpen}{Sharpen}

\DeclareMathOperator*{\expectation}{\mathbb{E}}

\newcommand{\graphmixkk}{\normalfont{\text{GraphMix}}}
\newcommand{\gnn}{\normalfont{\text{GNN}}}
\newcommand{\fcn}{\normalfont{\text{FCN}}}
\newcommand{\Dcal}{\mathcal{D}}
\newcommand{\Rcal}{\mathcal{R}}
\newcommand{\Fcal}{\mathcal{F}}
\newcommand{\Scal}{\mathcal{S}}
\newcommand{\Acal}{\mathcal{A}}
\newcommand{\Lcal}{\mathcal{L}}
\newcommand{\EE}{\mathbb{E}} 
\ifx\BlackBox\undefined
\newcommand{\BlackBox}{\rule{1.5ex}{1.5ex}}  
\fi
\ifx\proof\undefined
\newenvironment{proof}{\par\noindent{\emph{Proof.}}}{\hfill\BlackBox\\[2mm]}
\fi

\newtheorem{theorem}{Theorem}

\usepackage{cleveref}

\title{GraphMix: Improved Training of GNNs for Semi-Supervised Learning}

\author{Vikas Verma$^{1,2,5,\dagger}$, Meng Qu$^{1,2}$, Kenji Kawaguchi$^{4}$, Alex Lamb$^{1,2}$, Yoshua Bengio$^{1,2}$, \\ \textbf{Juho Kannala}$^{5}$, \textbf{Jian Tang}$^{1,3}$  \\
$^{1}$Mila - Qu\'ebec Artificial Intelligence Institute, Montr\'eal, Canada\\
$^{2}$Universit\'e de Montr\'eal, Canada \\
$^{3}$HEC, Montr\'eal, Canada\\
$^{4}$MIT, Cambridge,USA \\
$^{5}$Aalto University, Finland \\
$^{\dagger}$ \texttt{vikas.verma@aalto.fi}
}

\begin{document}

\maketitle

\begin{abstract}
  We present \textit{GraphMix}, a regularization method for Graph Neural Network based semi-supervised object classification, whereby we propose to train a fully-connected network jointly with the graph neural network via parameter sharing and interpolation-based regularization.
  Further, we provide a theoretical analysis of how \graphmix improves the generalization bounds of the underlying graph neural network, without making any assumptions about the "aggregation" layer or the depth of the graph neural networks. We experimentally validate this analysis by applying GraphMix to various architectures such as Graph Convolutional Networks, Graph Attention Networks and Graph-U-Net. Despite its simplicity, we demonstrate that GraphMix can consistently improve or closely match state-of-the-art performance using even simpler architectures such as Graph Convolutional Networks, across three established graph benchmarks:  Cora, Citeseer and Pubmed citation network datasets, as well as three newly proposed datasets: Cora-Full, Co-author-CS and Co-author-Physics. 
\end{abstract}

\section{Introduction}

Due to the presence of graph-structured data across a wide variety of domains, such as biological networks, citation networks and social networks, there have been several attempts to design neural networks, known as graph neural networks (GNN), that can process arbitrarily structured graphs. Early work includes \citep{gori2005new, scarselli2009} which propose a neural network that can directly process most types of graphs e.g., acyclic, cyclic, directed, and undirected graphs. More recent approaches include \citep{bruna, henaff, defferrard, kipf2016variational,  gilmer2017neural,  hamilton2017inductive, velivckovic2018graph, velickovic2018deep,  gmnn,  u-g-net, ma2019disentangled}, among others. Many of these approaches are designed for addressing the problem of  semi-supervised learning over graph-structured data \citep{gnn_review}.  Much of these research efforts have been dedicated to developing novel architectures.

Here we instead propose an architecture-agnostic method for regularized training of GNNs for semi-supervised node classification. Recently, regularization based on data-augmentation has been shown to be very effective in other types of neural networks but how to apply these techniques in GNNs is still under-explored. Our proposed method  \graphmix\footnote{code available at https://github.com/vikasverma1077/GraphMix} is 
a unified framework that draws inspiration from interpolation based data augmentation \citep{mixup, manifold_mixup} and  self-training based data-augmentation \citep{laine2016temporal, meanteacher,ict, mixmatch}.
We show that with our proposed method, we can achieve state-of-the-art performance even when using simpler GNN architectures such as Graph Convolutional Networks \citep{kipf2016semi}, with no additional memory cost and with minimal additional computation cost. Further, we conduct a theoritical analysis to demonstrate the effectiveness of the proposed method over the underlying GNNs.

\begin{figure*}[ht]
  \centering
  \includegraphics[width=0.9\textwidth]{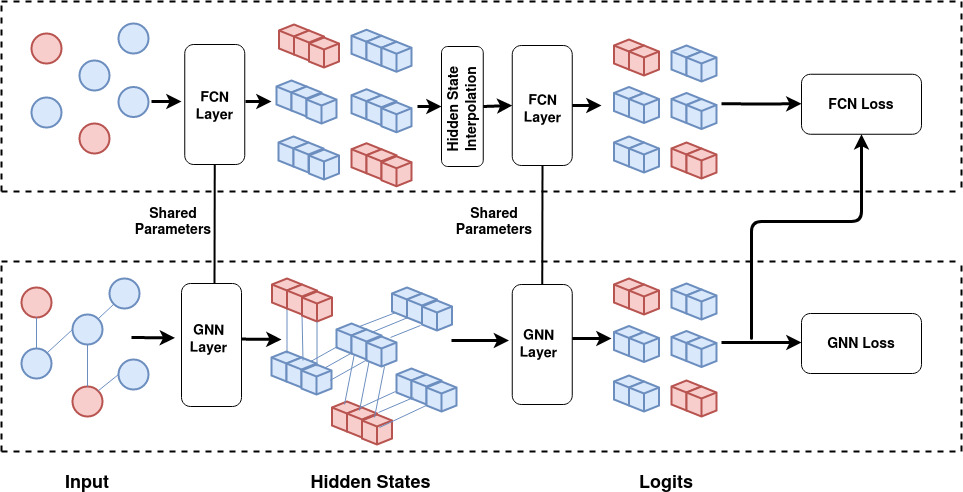}
  \caption{The procedure for training with \graphmix. The labeled and unlabeled nodes are shown with different colors in the graph. \graphmix augments the training of a baseline Graph Neural Network (GNN) with a Fully-Connected Network (FCN). The FCN is trained by interpolating the hidden states  and the corresponding labels. This leads to better features which are transferred to the GNN via sharing the linear transformation parameters $W$ (in Equation\ref{eq:gnn}) of the GNN and FCN layers. Furthermore, the predictions made by the GNN for unlabeled data are used to augment the input data for the FCN. The FCN and the GNN losses are minimized jointly by alternate minimization. }
  \label{fig::graphmix}
\end{figure*}

\section{Problem Definition and Preliminaries}
    
\paragraph{Problem Setup:}
We are interested in the problem of semi-supervised node and edge classification using graph-structured data. We can formally define such graph-structured data as an undirected graph ${\mathcal{G}} = ({\mathcal{V}}, \mathcal{A}, \mathcal{X})$, where $\mathcal{V}=\mathcal{V}_l\cup \mathcal{V}_u$ is the union of labeled ($\mathcal{V}_l$) and unlabeled ($\mathcal{V}_u$) nodes in the graph with cardinalities $n_l$ and $n_u$, and 
$\mathcal{A}$ is the adjacency matrix representing the edges between the nodes of $\mathcal{V}$,
$\mathcal{X} \in \mathbb{R}^{(n_l+n_u) \times d}$ is the input node features. Each node $v$ belongs to one out of $C$ classes and can be labeled with a $C$-dimensional one-hot vector ${y}_v\in \mathbb{R}^C$. Given the labels $Y_l \in \mathbb{R}^{n_l\times C} $ of the labeled nodes $\mathcal{V}_l$, the task is to predict the labels  $Y_{u}\in \mathbb{R}^{n_u\times C}$ of the unlabeled nodes  $\mathcal{V}_{u}$.

\paragraph{Graph Neural Networks:}
Graph Neural Networks (GNN) learn the $l_{th}$ layer representations of a sample $i$ by leveraging the  representations of the samples $NB(i)$  in the neighbourhood of $i$. This is done by using an aggregation function that takes as an input the representations of all the samples along with the graph structure and outputs the aggregated representation. The aggregation function can be defined using the Graph Convolution layer \citep{kipf2016semi}, Graph Attention Layer \citep{velivckovic2018graph}, or any general message passing layer \citep{gilmer2017neural}. Formally, let  $\mathbf{h}^{(l)} \in \mathbb{R}^{n \times k}$ be a matrix containing the  $k$-dimensional representation of $n$ nodes in the $l_{th}$ layer, then:
\begin{align}
\label{eq:gnn}
    \mathbf{h}^{(l+1)} = \sigma(AGGREGATE(\mathbf{h}^{(l)} \mathbf{W}, \mathcal{A} ))
\end{align}

where $\mathbf{W} \in \mathbb{R}^{k \times k'}$ is a linear transformation matrix, $k'$ is the dimension of $(l+1)_{th}$ layer, $AGGREGATE$ is the aggregation function that utilizes the graph adjacency matrix $\mathcal{A}$ to aggregate the hidden representations of neighbouring nodes and $\sigma$ is a non-linear activation function, e.g. ReLU.

\paragraph{Interpolation Based Regularization Techniques:}
\label{subsec:mm}
Recently, interpolation-based techniques have been proposed for regularizing neural networks.  We briefly describe some of these techniques here. Mixup \citep{mixup} trains a neural network on the convex combination of input and targets, whereas Manifold Mixup \citep{manifold_mixup} trains a neural network on the convex combination of the hidden states  of a randomly chosen hidden layer and the targets. While Mixup regularizes a neural network by enforcing that the model output should change linearly in between the examples in the input space, Manifold Mixup regularizes the neural network by learning better (more discriminative) hidden states.

Formally, suppose $ T_{\theta}(\mathbf{x})=(f\circ g)_{\theta} (\mathbf{x})$ is a neural network parametrized with $\theta$ such that $g:\mathbf{x}\xrightarrow{}\mathbf{h}$ is a function that maps input sample to hidden states, $f:\mathbf{h}\xrightarrow{}\mathbf{\hat{y}}$ is a function that maps hidden states to predicted output, $\lambda$ is a random variable drawn from $\betadist(\alpha, \alpha)$ distribution,  $\text{Mix}_{\lambda}(\mathbf{a},\mathbf{b}) =\lambda*\mathbf{a}+(1-\lambda)*\mathbf{b}$ is an interpolation function, $\mathcal{D}$ is the data distribution, $(\mathbf{x},\mathbf{y})$ and $(\mathbf{x'},\mathbf{y'})$ is a pair of labeled examples sampled from distribution $\mathcal{D}$ and $\ell$ be a loss function such as cross-entropy loss, then the Manifold Mixup Loss is defined as:

\begin{align}
\label{eq:mixup_form}
    \mathcal{L}_{\text{MM}}(\mathcal{D},T_{\theta}, \alpha) = &\expectation_{(\mathbf{x},\mathbf{y}) \sim \mathcal{D}}\,
    \expectation_{(\mathbf{x'}, \mathbf{y'}) \sim \mathcal{D}}\,
    \expectation_{\lambda \sim \text{Beta}(\alpha, \alpha)}\, 
     &\ell(f(\text{Mix}_\lambda(g(\mathbf{x}), g(\mathbf{x'}))), \text{Mix}_\lambda(\mathbf{y}, \mathbf{y'})).
\end{align}

We use above Manifold Mixup loss for training an auxiliary Fully-connected-network as described in Section \ref{method}.


\section{GraphMix}
\label{method}

\subsection{Motivation} Data Augmentation is one of the simplest and most efficient technique for regularizing a neural network. In the domains of computer vision, speech and natural language, there exist efficient data augmentation techniques, for example, random cropping, translation or Cutout \citep{cutout} for computer vision, \citep{audioaf} and \citep{specaugment} for speech and \citep{dataNoising} for natural language. However, data augmentation for the graph-structured data remains under-explored. There exists some recent work along these lines but the prohibitive computation cost (see Section \ref{sec:relatedwork_reg}) introduced by these methods make them impractical for real-world large graph datasets. Based on these limitations, our main objective is to propose an efficient data augmentation technique for graph datasets. 

Recent work based on interpolation-based data augmentation  \citep{mixup,manifold_mixup} has seen sizable improvements in regularization performance across a number of tasks.  However, these techniques are not directly applicable to graphs for an important reason:\textit{ Although we can create additional nodes by interpolating the features and corresponding labels, it remains unclear how these new nodes must be connected to the original nodes via synthetic edges such that the structure of the whole graph is preserved.} To alleviate this issue, we propose to train an auxiliary Fully-Connected Network (FCN) using Manifold Mixup as discussed in Section \ref{sec:method}. Note that the FCN only uses the node features (not the graph structure), thus the Manifold mixup loss in Eq. \ref{eq:mixup_form} can be directly used for training the FCN.

Interpolation based data-augmentation techniques have an added advantage for training GNNs:  A vanilla GNN learns the representation of each node by iteratively aggregating information from the neighbors of that node (Equation ~\ref{eq:gnn}). However, this induces the problem of \textit{oversmoothing}  \citep{Li2018DeeperII,pmlr-v80-xu18c} while training GNNs with many layers. Due to this limitation,  GNNs are trained only with a few layers, and thus they can only leverage the local neighbourhood of each node for learning its representations, without leveraging the representations of the nodes which are multiple hops away in the graph. This limitation can be addressed using the interpolation-based method such as Manifold Mixup: in Manifold Mixup, the representations of a randomly chosen pair of nodes is used to facilitate better representation learning; it is possible that the randomly chosen pair of nodes will not be in the local neighbourhood of each other. 
Based on these challenges and motivations we present our proposed approach \graphmix for training GNNs in the following Section.

\subsection{Method}
\label{sec:method}
  We first describe \graphmix at a high-level and then give a more formal description.  \graphmix augments the vanilla GNN with a Fully-Connected Network (FCN). The FCN loss is computed using \mm as discussed below and the GNN loss is computed normally.  \mm training of FCN facilitates learning more discriminative node representations \citep{manifold_mixup}. An important question is how these more discriminative node representations can be transferred to the GNN?  One potential approach could involve maximizing the mutual information between the hidden states of the FCN and the GNN using formulations similar to those proposed by \citep{infomax,infoGraph}.  However, this requires optimizing additional network parameters. Instead, we propose parameter sharing between FCN and GNN to facilitate the transfer of discriminative node representations from the FCN to the GNN. It is a viable option because as mentioned in Eq \ref{eq:gnn}, a GNN layer typically performs an additional operation ($AGGREGATE$) on the linear transformations of node representations (which are essentially pre-activation representations of the FCN layer). Using the more discriminative representations of the nodes from FCN, as well as the graph structure, the GNN loss is computed in the usual way to further refine the node representations.  In this way we can exploit the improved representations from \mm for training GNNs.
  
  
 In Section \ref{analysis}, without making any assumption about the aggregation function and the depth of the graph neural network, we show that GraphMix improves the generalization of the underlying graph neural network. This makes GraphMix applicable to various kind of architectures having different aggregation functions, such as weighed averaging in GCN \citep{kipf2016variational}, attention based aggregation in GAT \citep{velivckovic2018graph} and graph-pooling/unpooling operations in Graph U-Nets \citep{u-g-net}. 
 In the aforementioned sense, \graphmix procedure is highly flexible: it can be applied to any underlying GNN as long as the underlying GNN applies parametric transformations to the node features.
 
 Additionally, we propose to use the predicted targets from the GNN to augment the training set of the FCN. In this way, both the FCN and the GNN facilitate each other's learning process. Both the FCN loss and the GNN loss are optimized in an alternating fashion during training.  At inference time,  predictions are made using only the GNN. 
  
   A diagram illustrating \graphmix is presented in Figure~\ref{fig::graphmix} and the full algorithm is presented in Appendix \ref{app:algo}. Further, we draw similarities and difference of \graphmix w.r.t. Co-training framework in the Appendix \ref{sec:cotraining}.

So far we have presented the general design of \graphmixns, now we present \graphmix more formally. Given a graph $\mathcal{G}$, let $(\mathbf{X}_l,\mathbf{Y}_l) $ be the input features and the labels of the labeled nodes $\mathcal{V}_l$ and let $(\mathbf{X}_u)$ be the input features of the unlabeled nodes $\mathcal{V}_u$. Let $F_{\theta}$ and $G_{\theta}$ be a FCN and a GNN respectively, which share the parameters $\theta$. The FCN loss from the labeled data is computed using Eq. \ref{eq:mixup_form} as follows:
\begin{align}
\label{eq:fcn_sup}
    \mathcal{L}_{\text{supervised}} = \mathcal{L}_{\text{MM}}((\mathbf{X}_l,\mathbf{Y}_l),F_{\theta},\alpha)
\end{align}
For unlabeled nodes $\mathcal{V}_u$, we compute the prediction $\mathbf{\hat{Y}_u}$ using the GNN:
\begin{align}
\label{eq:predict}
\mathbf{\hat{Y}_u} = G_{\theta}(\mathbf{X}_u)
\end{align}
We note that recent state-of-the-art semi-supervised learning methods use a \textit{teacher} model to accurately predict targets for the unlabeled data. The teacher model can be realized as a temporal ensemble of the \textit{student} model (the model being trained) \citep{laine2016temporal} or by using an Exponential Moving Average (EMA) of the parameters of the student model \citep{meanteacher, ict}. Different from these approaches, we use the GNN as a teacher model for predicting the targets for the FCN. This is due to the fact that GNNs leverage graph structure, which in practice, allows them to make more accurate predictions than the temporal ensemble or EMA ensemble of FCN (although there is no theoretical guarantee for this).

Moreover, to improve the accuracy of the predicted targets in Eq \ref{eq:predict}, we applied the average of the model prediction on $K$ random perturbations of an input sample along with sharpening as discussed in Appendix ~\ref{sec::predict}.

Using the predicted targets for unlabeled nodes, we create a new training set $(\mathbf{X}_u,\mathbf{\hat{Y}_u})$. The loss from the unlabeled data for the FCN is computed as:
\begin{align}
\label{eq:fcn_usup}
    \mathcal{L}_{\text{unsupervised}} = \mathcal{L}_{\text{MM}}((\mathbf{X}_u,\mathbf{\hat{Y}}_u),F_{\theta},\alpha)
\end{align}
Total loss for training the FCN is given as the weighted sum of above two loss terms.
\begin{align}
\label{eq:fcn_loss}
    \mathcal{L}_{\text{FCN}}= \mathcal{L}_{\text{supervised}} + w(t)*\mathcal{L}_{\text{unsupervised}}
\end{align}
 where $w(t)$ is a sigmoid ramp-up function \citep{meanteacher} which increases from zero to its max value $\gamma$ during the course of training.

Now let us assume that the loss for an underlying GNN is $\mathcal{L}_{\text{GNN}}=\ell(G_{\theta}(\mathbf{X}_l), \mathbf{Y}_l)$; the overall \graphmix loss for the joint training of the FCN and GNN can be defined as the weighted sum of the GNN loss and the FCN loss:
\begin{align}
\label{eq:graphmix_loss}
    \mathcal{L}_{\text{\graphmix}}= \mathcal{L}_{\text{GNN}} + \lambda*\mathcal{L}_{\text{FCN}}
\end{align}

However, throughout our experiments, optimizing FCN loss and GNN loss alternatively at each training epoch achieved better test accuracy (discussed in Appendix \ref{app:joint_alternate}). This has an added benefit that it removes the need to tune weighing hyper-parameter $\lambda$.

For \mm training of FCN, we apply \textit{mixup} only in the hidden layer. Note that in \citep{manifold_mixup}, the authors recommended applying mixing in a randomly chosen layer (which also includes the input layer) at each training update.  However, we observed under-fitting when applying \textit{mixup} randomly at the input layer or the hidden layer. Applying \textit{mixup} only in the input layer also resulted in underfitting and did not improve test accuracy.

\textbf{Memory and Computational Requirements:} One of the major limitations of current GNNs, which prohibits their application to real-world large datasets, is their memory complexity. For example, the fastest implementation of GCN, which stores the entire adjacency matrix $\mathcal{A}$ in the memory, has the memory complexity $O(|\mathcal{V}|^2)$. Implementations with lower memory requirements are possible but they incur higher latency cost due to repeatedly loading parts of adjacency matrix in the memory. Due to these reasons, methods which have additional memory requirement in comparison to the baseline GNNs, are less appealing in practice. In \graphmixns, since the parameters of the FCN and GNN are shared, there is no additional memory requirement. Furthermore, \graphmix does not add any \textit{significant computation cost} over the underlying GNN, because the underlying GNN is trained in the standard way and the FCN training requires trivial additional computation cost for computing the predicted-targets (Appendix \ref{sec::predict}) and the interpolation function ( $\text{Mix}_{\lambda}(\mathbf{a},\mathbf{b})$ in Section \ref{subsec:mm}).

\subsection{Analysis}
\label{analysis}
In this subsection, we study how GraphMix impacts the generalization bound of a underlying GNN.  Our analysis, which is based on Rademacher complexity \citep{bartlett2002rademacher}, provides a new generalization bound for GraphMix, which shows how adding regularization to training the FCN with pseudolabels improves overall generalization. We rely on the effect of changing one sample in Manifold Mixup and the fact that the weights are shared by a GNN and the corresponding FCN in GraphMix. 

Let $G$ be a fixed graph with $n$ total nodes. Define $z_{i}=(x_i, y_i)$ to be the  pair of the feature and the true label of the $i$-th node. Without loss of generality, let $S=(z_{1},\dots,z_m)$ be the training set with the labeled nodes where  $m=n_l$ and data points $z_{1},\dots,z_m$ are sampled according to an unknown distribution $\Dcal$ to form the labeled node set $S$. In this subsection, we follow the previous paper on graph neural networks by assuming that  all samples are  i.i.d. (including  replacement sample)  \citep{verma2019stability}.

Let $\Gamma$ be a finite set of the hyperparameters. For every hyperparameter $\gamma \in \Gamma$, define $\Fcal_\gamma$ to be a distribution-dependent hypothesis space corresponding the hyperparameter $\gamma$. That is, $\Fcal_\gamma=\{f_\gamma :(\exists S\in \Scal) [f_\gamma =\Acal_{\gamma}(S)]\}$ where $\Acal_{\gamma}$ is an algorithm that outputs the hypothesis $f_\gamma$ given a dataset $S$, and $\Scal$ is the set of training datasets such that the probability of $S\in \Scal$ according to  $\Dcal$ is one. For each  $f_\gamma \in \Fcal_\gamma$,  $f_\gamma(\cdot \ ; G)$ represents GNN with the graph $G$ and $f_\gamma(\cdot \ ; G_{0})$ represents FCN\ where $G_0$ is  the null graph version of $G$ (i.e., $G$ without edges).  Let $\Rcal_{m}^{\ell}(\Fcal_\gamma)$ be the Rademacher complexity \citep{bartlett2002rademacher} of the set $\{(x,y) \mapsto\ell(f_\gamma(x;G),y):f_\gamma \in\Fcal_\gamma \}$.

Let $L_{\gnn}(S,f_\gamma)$ be the $\Lcal_{\gnn}$ with the GNN $f_\gamma(\cdot \ ; G)$ and labeled data points $S$ as $L_{\gnn}(S,f_\gamma)=\frac{1}{m}\sum_{i=1}^m \ell(f_\gamma(x_{i};G),y_{{i}})$. Let $L_{\fcn}(S,f_\gamma)$ be $\Lcal_{\fcn}$ with the FCN $f_\gamma(\cdot \ ; G_{0})$ and labeled data points $S$ as  $L_{\fcn}(S,f_\gamma)= \Lcal_{\text{MM}}(S,f_\gamma(\cdot \ ; G_{0}),\alpha)+\frac{n_u}{m} \Lcal_{\text{MM}}((\mathbf X_{u}^S, \mathbf {\hat Y}_u^S),f_\gamma(\cdot \ ; G_{0}),\alpha)$ where $(\mathbf X_{u}^S, \mathbf {\hat Y}_u^S)$ is the unlabeled nodes $(\mathbf X_{u}, \mathbf {\hat Y}_u)$ that corresponds to the labeled node set $S$. Let $L_{\graphmixkk}(S,f_\gamma)=L_{\gnn}(S,f_\gamma)+\lambda L_{\fcn}(S,f_\gamma)$. Let $c$ be the upper bound on per-sample loss as $c\ge\ell(f_\gamma(x_{i};G),y_{{i}})$ and $c \ge\ell(f_\gamma(x_{i};G_0),y_{{i}})$. For example, $c=1$ for training and test error (or  0-1 loss).

Theorem \ref{thm:gen} provides a generalization bound for GraphMix, which shows that GraphMix can improve the generalization bound of the underlying GNN under the condition of $V>0$ as discussed below.

\begin{theorem} \label{thm:gen}
For any $\delta>0$, with probability at least $1-\delta$, the following holds: for all $\gamma \in \Gamma$ and all $f_\gamma \in \Fcal_\gamma$, we have that $\EE_{x,y \sim\Dcal}[\ell(f_\gamma(x;G),y)] -L_{\graphmixkk}(S,f_\gamma)
\le \Rcal_{m}^{\ell}(\Fcal_\gamma) +c \sqrt{\frac{\ln(|\Gamma|/\delta)}{2m}} - \lambda V$,
where 
$
V= \EE_{S' \sim \Dcal^m} [ \inf_{f_\gamma \in \Fcal_\gamma} L_{\fcn}(S',f_\gamma) ] - 4c\sqrt{\frac{\ln(|\Gamma|/\delta)}{2m}}. 
$ 
\end{theorem}
The proof of Theorem \ref{thm:gen} is given in the following. The generalization bound in Theorem \ref{thm:gen} becomes a  generalization bound for GNN without GraphMix when $\lambda=0$ as desired. By comparing the generalization bound with $\lambda =0$ (no GraphMix) and $\lambda >0$ (with GraphMix), we can see that GraphMix  improves the generalization bound of underlying GNN when 
$
V= \EE_{S' \sim \Dcal^m} [ \inf_{f_\gamma \in \Fcal_\gamma} L_{\fcn}(S',f_\gamma) ] - 4c\sqrt{\ln(|\Gamma|/\delta)/(2m)} > 0.
$  
Here, the first term $ \EE_{S' \sim \Dcal^m} [ \inf_{f_\gamma \in \Fcal_\gamma} L_{\fcn}(S',f_\gamma) ]$ increases as the hypothesis space   $\Fcal_\gamma$ gets ``smaller'' or has  less complexity.
Thus, GraphMix improves the generalization bound of an underlying GNN when   the hyperparameter search over $\gamma\in \Gamma$ results in  $\Fcal_\gamma$ of moderate complexity such that the first term is greater than $ 4c\sqrt{\ln(|\Gamma|/\delta)/(2m)}$. The first term contains the manifold mixup loss $L_{\fcn}(S',f_\gamma)$ over random training datasets $S'$ ($\neq S$), which is expected to be larger when compared with that of the standard loss without manifold mixup.  

For each fixed $\Fcal_\gamma$, the generalization bound in Theorem \ref{thm:gen} goes to zero as $m \rightarrow \infty$ since $\Rcal_{m}^{\ell}(\Fcal_\gamma) \rightarrow 0$ and $V \rightarrow V_{0}\ge0 $ as $m \rightarrow \infty$. The training loss is also minimized when the trained model $f_{\gamma} \in \Fcal_\gamma$  fits well to the particular given training data set $S$.   Therefore, given a particular training dataset $S$, the expected loss is minimized if we conduct a hyperparameter search over  $\gamma\in \Gamma$ such that $f_\gamma \in \Fcal_\gamma$  minimize the training loss for the given $S$ but $\Fcal_\gamma$ has moderate complexity to not being able to minimize the  manifold mixup losses for other datasets $S'$ ($\neq S$) drawn from $\Dcal$. Unlike the standard data points, the data points generated during  manifold mixup in GraphMix are typically not memorizable or interpolated by FCN.

\subsection{Proof of Theorem \ref{thm:gen}} \label{sec:app:proof:gen}
In the proof, we analyse the effect of changing one sample in Manifold Mixup and GNN-generated targets by utilizing the fact that the weights are shared by a GNN and the corresponding FCN. The the fact of sharing weights also results in the generalization bound that relates the generalization in FCN via Manifold Mixup to the generalization of the GNN.

\begin{proof}
Let $\gamma \in \Gamma$ be fixed. Define $\varphi(S)= \sup_{f_\gamma \in \Fcal_\gamma} \EE_{x,y \sim\Dcal}[\ell(f_\gamma(x;G),y)]-L_{\graphmixkk}(S,f_\gamma)$.
We first provide an upper bound on $\varphi(S)$ by using McDiarmid's inequality. To apply McDiarmid's inequality to $\varphi(\Dcal)$, we compute an upper bound on $|\varphi(S)-\varphi(S')|$ where  $S$ and $S'$ be two training datasets differing by exactly one point of an arbitrary index $i_{0}$; i.e.,  $S_i= S'_i$ for all $i\neq i_{0}$ and $S_{i_{0}} \neq S'_{i_{0}}$. 
\begin{align*}
\varphi(S')-\varphi(S) 
 &\le \sup_{f_\gamma \in \Fcal_\gamma} L_{\graphmixkk}(S,f_\gamma)-L_{\graphmixkk}(S',f_\gamma) \\ &=  \sup_{f_\gamma \in \Fcal_\gamma} (L_{\gnn}(S,f_\gamma)-L_{\gnn}(S',f_\gamma)) \\ & +\lambda(L_{\fcn}(S,f_\gamma)-L_{\fcn}(S',f_\gamma))
\end{align*}
where the first line follows the property of the supremum, $\sup (a) - \sup (b)\le\sup(a-b)$, and the second line follows the definition of $L_{\graphmixkk}$.

For the first term,
\begin{align*}
L_{\gnn}(S,f_\gamma)-L_{\gnn}(S',f_\gamma)
 = \\ \frac{1}{m}\left(\ell(f_\gamma(x_{i_0};G),y_{{i_0}})-  (\ell(f_\gamma(x_{i_0}';G),y_{{i_0}}') \right) \le \frac{c}{m}, 
\end{align*}
where we used the fact that given a  fixed $G$ and a fixed $f_\gamma$, $\ell(f_\gamma(x_{i};G),y_{{i}})=\ell(f_\gamma(x_{i}';G),y_{{i}}')$ for $i\neq i_0$. This holds since  $f_\gamma(\cdot \ ;G)$ does not depend on $S$ given a   $G$ and a  $f_\gamma$,
  even though $f_\gamma(x_{i};G)$ contains the aggregation functions over the graph $G$. 

For the second term, 
\begin{align*}
\scalebox{1.0}{$\displaystyle \Lcal_{\text{MM}}(S,f_\gamma(\cdot \ ; G_{0}),\alpha) - \Lcal_{\text{MM}}(S',f_\gamma(\cdot \ ; G_{0}),\alpha) $} \\ 
\le \frac{c (2m-1)}{m^2} 
\le \frac{2c}{m},
 \end{align*}
where we use the fact that $\Lcal_{\text{MM}}(S,f_\gamma(\cdot \ ; G_{0}),\alpha)$ has $m^2$ terms and $2m-1$ terms differ for $S$ and $S'$, each of which is bounded by the constant $c$. Similarly, 
$
\Lcal_{\text{MM}}((\mathbf X_{u}^S, \mathbf {\hat Y}_u^S),f_\gamma(\cdot \ ; G_{0}),\alpha)-\Lcal_{\text{MM}}((\mathbf X_{u}^{S'}, \mathbf {\hat Y}_u^{S'}),f_\gamma(\cdot \ ; G_{0}),\alpha) \le \frac{2c}{n_u} 
$, since the labels $\mathbf {\hat Y}_u^S$ and $\mathbf {\hat Y}_u^{S'}$ are determined by $f_\gamma(x_{i};G)$, and $f_\gamma(x_{i};G)=f_\gamma(x_{i}';G)$ for $i\neq i_0$, given a  fixed $G$ and a fixed $f_\gamma$. Therefore,
\begin{align*}
L_{\fcn}(S,f_\gamma)-L_{\fcn}(S',f_\gamma) \le \frac{4c}{m}. 
\end{align*}
Using these upper bounds, 
\begin{align*}
\varphi(S')-\varphi(S) \le \frac{c(1+4 \lambda)}{m}.  
\end{align*}
Similarly, $\varphi(S)-\varphi(S') \le \frac{c(1+4 \lambda)}{m}$. Thus, by McDiarmid's inequality, for any $\delta>0$, with probability at least $1-\delta/ |\Gamma|$,
$$
\varphi(S) \le  \EE_{\bar S}[\varphi(\bar S)] + c(1+4 \lambda) \sqrt{\frac{\ln( |\Gamma|/\delta)}{2m}}.
$$
Moreover, 
\begin{align*}
\EE_{\bar S}[\varphi(\bar S)] + \lambda\EE_{\bar S}\left[\inf_{f_\gamma \in \Fcal_\gamma}  L_{\fcn}(\bar S,f_\gamma) \right] \\
  \le \EE_{\bar S}\left[\sup_{f_\gamma \in \Fcal_\gamma} \EE_{\bar x',\bar y' \sim\Dcal}[\ell(f_\gamma(\bar x';G), \bar y')]-L_{\gnn}(\bar S,f_\gamma)\right]   
 \\  \scalebox{0.95}{$\displaystyle \le\EE_{\bar S,\bar S'}\left[\sup_{f_\gamma \in \Fcal_\gamma} \frac{1}{m}\sum_{i=1}^m (\ell(f_\gamma(\bar x_{i}';G), \bar y_{i}')-\ell(f_\gamma(\bar x_i;G),\bar  y_{i}))\right] $}  
 \\  \scalebox{0.9}{$\displaystyle \le\EE_{\xi, \bar \Dcal, \bar \Dcal'}\left[\sup_{f_\gamma \in \Fcal_\gamma} \frac{1}{m}\sum_{i=1}^m  \xi_i(\ell(f_\gamma(\bar x_{i}';G), \bar y_{i}')-\ell(f_\gamma(\bar x_i;G),\bar y_{i}))\right]  $} 
  \\  \le 2\Rcal_{n}(\Theta)    
\end{align*}
where the second line and  the last line use the subadditivity of supremum, the third line uses the Jensen's inequality and the convexity of  the 
supremum, the fourth line follows that for each $\xi_i \in \{-1,+1\}$, the distribution of each term $\xi_i (\ell(f_\gamma(\bar x_{i}';G),\bar y_{i}')-\ell(f_\gamma(\bar x_i;G),\bar y_{i}))$ is the  distribution of  $(\ell(f_\gamma(\bar x_{i}';G),\bar y_{i}')-\ell(f_\gamma(\bar x_i;G),\bar y_{i}))$  since $\bar S$ and $\bar S'$ are drawn iid with the same distribution.

Therefore, for any $\delta>0$, with probability at least $1-\delta/|\Gamma|$,
all $f_\gamma \in \Fcal_\gamma$,
\begin{align*}
\EE_{x,y \sim\Dcal}[\ell(f_\gamma(x;G),y)]-L_{\graphmixkk}(S,f_\gamma) \\
 \le \Rcal_{m}^{\ell}(\Fcal_\gamma) +c \sqrt{\frac{\ln(|\Gamma|/\delta)}{2m}} - \lambda V. 
\end{align*}
by taking union bounds over all $\gamma \in \Gamma$, we obtain the statement of this theorem.
 
\end{proof}

\section{Experiments}
\label{exp}

We present results for \graphmix algorithm using standard benchmark datasets and the standard architecture in Section \ref{subsec:results} and \ref{linkclass}. We also conduct an ablation study on \graphmix in Section \ref{subsection:ablation} to understand the contribution of various components to its performance. Refer to Appendix~\ref{app:datasets} for dataset details and   \ref{app:hyper} for  implementation and hyperparameter details.

\subsection{Semi-supervised Node Classification}
\label{subsec:results}

 For baselines, we choose GCN \citep{kipf2016semi}, and the recent state-of-the-art graph neural networks GAT \citep{velivckovic2018graph}, GMNN \citep{gmnn} and Graph U-Net \citep{u-g-net}, as well as the recently proposed regularizers for graph neural networks.
 \graphmixgcn,  \graphmixgat and GraphMix(Graph U-Net) refer to the methods where underlying GNNs are GCN, GAT and Graph U-Net respectively. Refer to Appendix \ref{app:hyper} for implementation and hyperparameter details.
\citep{pitfalls} demonstrated that the performance of the current state-of-the-art Graph Neural Networks on the standard train/validation/test split of the popular benchmark datasets (such as Cora, Citeseer, Pubmed, etc) is significantly different from their performance on the random splits. For fair evaluation, they recommend using multiple random partitions of the datasets. Along these lines, we created $10$ random splits of the Cora, Citeseer and Pubmed with the same train/ validation/test number of samples as in the standard split. We also provide the results for the standard train/validation/test split.
%
The results are presented in Table \ref{tab::random_split}. We observe that GraphMix always improves the accuracy of the underlying GNNs such as GCN, GAT and Graph-U-Net across all the dataset, with \graphmixgcn achieving the best results.
We further present results with fewer labeled samples in Appendix \ref{sec:fewerlabels}. We observer that the relative increase in test accuracy using \graphmix over the baseline GNN is more pronounced when the labeled samples are fewer. This makes \graphmix particularly appealing for very few labeled data problems.

\begin{table}
           \centering
           \captionsetup[subtable]{position = top}
           \captionsetup[table]{position=top}
           \caption{Results of node classification (\% test accuracy) on the standard split of datasets. [*] means the results are taken from the corresponding papers. We conduct 100 trials and report mean and standard deviation over the trials (refer to Table \ref{tab:full} in the Appendix for comparison with other methods on standard Train/Validation/Test split). } 
	        \label{tab::standard_split}
               \resizebox{0.8\linewidth}{!}{
               		\begin{tabular}{c c c c}\hline
		        
		    	\textbf{Algorithm}	& \textbf{Cora} & \textbf{Citeseer} & \textbf{Pubmed} \\ 
		    	\midrule
		     GCN & 81.30$\pm$0.66  & 70.61$\pm$0.22 & 79.86$\pm$0.34\\
             GAT & 82.70$\pm$0.21  & 70.40$\pm$0.35 & 79.05$\pm$0.64 \\
             
             Graph U-Net & 81.74$\pm$0.54  & 67.69$\pm$1.10 & 77.73 $\pm$0.98 \\
             
             \midrule 
             BVAT*  & 83.6$\pm$0.5  &           74.0$\pm$0.6    & 79.9$\pm$0.4 \\
            DropEdge*    &    82.8 &            72.3     &             79.6 \\  
            GraphSGAN* &    83.0$\pm$1.3              &   73.1$\pm$1.8        &   - \\
            GraphAT*     &  82.5 &               73.4        &   -  \\
            GraphVAT*       & 82.6 & 73.7        & - \\
            \midrule
             \graphmix(GCN) & \textbf{83.94$\pm$0.57}  & \textbf{74.72$\pm$0.59} & 80.98$\pm$0.55 \\ 
             \graphmix(GAT) & 83.32$\pm$0.18  & 73.08$\pm$0.23 & \textbf{81.10$\pm$0.78} \\
             
             \graphmix(Graph U-Net) & 82.47$\pm$0.76 & 69.31$\pm$1.52 & 78.91$\pm$1.25  \\
		     \bottomrule

	        \end{tabular}
               }
           
\end{table}
\begin{table}
           \centering
           \captionsetup[subtable]{position = top}
           \captionsetup[table]{position=top}
           \caption{Results of node classification (\% test accuracy) using 10 random  Train/Validation/Test split of datasets. We conduct 100 trials and report mean and standard deviation over the trials. } 
	        \label{tab::random_split}
	           \resizebox{0.8\linewidth}{!}{
               		\begin{tabular}{c c c c}\hline
		        
		    	\textbf{Algorithm}	& \textbf{Cora} & \textbf{Citeseer} & \textbf{Pubmed} \\ 
		    	\midrule
		    	
		     GCN & 77.84$\pm$1.45  & 72.56$\pm$2.46 & 78.74$\pm$0.99\\
		     
		     GAT & 77.74$\pm$1.86  & 70.41$\pm$1.81  & 78.48$\pm$0.96 \\
		     
		     Graph U-Net & 77.59$\pm$1.60 & 67.55$\pm$0.69  &76.79$\pm$2.45 \\
		     
		     

		     \midrule
		     \graphmix(GCN) & \textbf{82.07$\pm$1.17}  & \textbf{76.45$\pm$1.57} &  \textbf{80.72$\pm$1.08}\\
		     
		     \graphmix(GAT) & 80.63$\pm$1.31  & 74.08$\pm$1.26  & 80.14$\pm$1.51 \\
		     \makecell{\graphmix(Graph-U-Net)} & 80.18$\pm$1.62   & 72.85$\pm$1.71  &78.47$\pm$0.64  \\

		     \bottomrule

	           \end{tabular}
               }
               
          
\end{table}


\subsection{Results on Larger Datasets}
\label{app:larger_datasets}

\begin{table}
\centering
\caption{Comparison of GraphMix with other methods (\% test accuracy ), for Cora-Full, Coauthor-CS, Coauthor-Physics. $*$ refers to the results reported in \citep{pitfalls}.}
\label{tab:result_new_dataset}
\resizebox{0.8\columnwidth}{!}{\begin{tabular}{l l l l}
\toprule
{\bf Algorithm} & {\bf Cora-Full} & {\bf \makecell[l]{Coauthor-CS}} & {\bf \makecell[l]{Coauthor \\-Physics}}\\ \midrule
GCN* & 62.2$\pm$0.6  & 91.1$\pm$0.5 & 92.8$\pm$1.0 \\
GAT* &  51.9$\pm$1.5 & 90.5$\pm$0.6 & 92.5$\pm$0.9 \\
MoNet* &  59.8$\pm$0.8 & 90.8$\pm$0.6 & 92.5$\pm$0.9 \\
GS-Mean* & 58.6$\pm$1.6 & 91.3$\pm$2.8 & 93.0$\pm$0.8 \\

 \midrule 
 GCN & 60.13$\pm$0.57 & 91.27$\pm$0.56 & 92.90$\pm$0.92 \\
 Graph-U-Net   & 59.82$\pm$0.39  &           90.89$\pm$0.43    & 92.57$\pm$0.81 \\
 \midrule
 \makecell[l]{\graphmix (GCN)} & \textbf{61.80$\pm$0.54} & \textbf{91.83$\pm$0.51} & \textbf{94.49$\pm$0.84} \\
 GraphMix (Graph-U-Net) &    60.92 $\pm$ 0.51 & 91.44  $\pm$ 0.46   &             93.78 $\pm$ 0.79 \\  
 
 \bottomrule
\end{tabular}}
\end{table}
We also provide results on three recently proposed datasets which are relatively larger than standard benchmark datasets (Cora/Citeseer/Pubmed). We use Cora-Full dataset proposed in  \citep{bojchevski2018deep} and Coauthor-CS and Coauthor-Physics datasets proposed in \citep{pitfalls}. The results are presented in Table \ref{tab:result_new_dataset}\footnote{We do not provide results for GAT based experiments in Table \ref{tab:result_new_dataset} and Table \ref{tab::result_link_classification} because we ran out of GPU memory required to run  these experiments with larger (higher number of nodes) datasets.}. We observe that \graphmixgcn improves the results over GCN for all the three datasets with a significant margin. We note that we did minimal hyperparameter search for \graphmixgcn as mentioned in Appendix \ref{app:hyper_largerdata}. The details of the datasets is given in Appendix \ref{app:datasets}.

\begin{figure*}[h!]
\centering
    \subfloat[GCN]{%
        \includegraphics[width=0.24\linewidth,trim={0cm 0cm 1.5cm 1.5cm},clip]{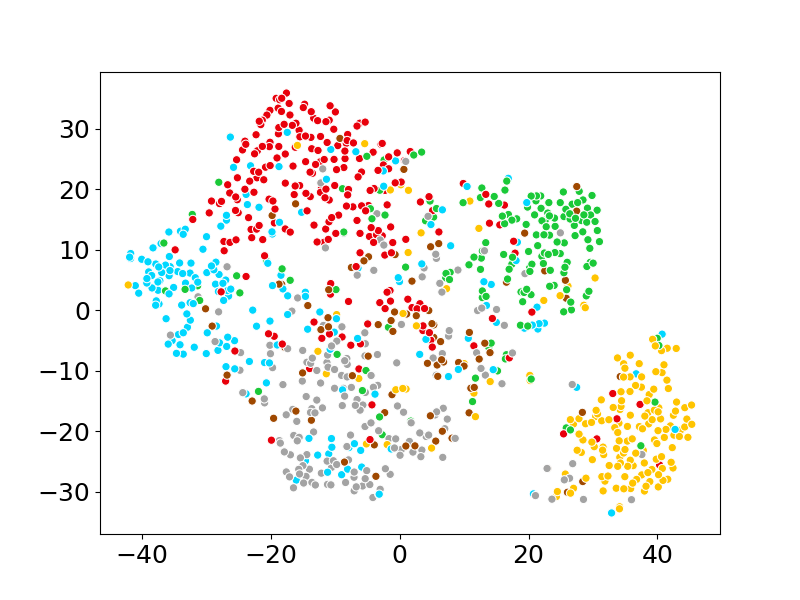}
        \label{fig:visualize_a}
    }
    \subfloat[GCN(self-training)]{%
        \includegraphics[width=0.24\linewidth,trim={0cm 0cm 1.5cm  1.5cm},clip]{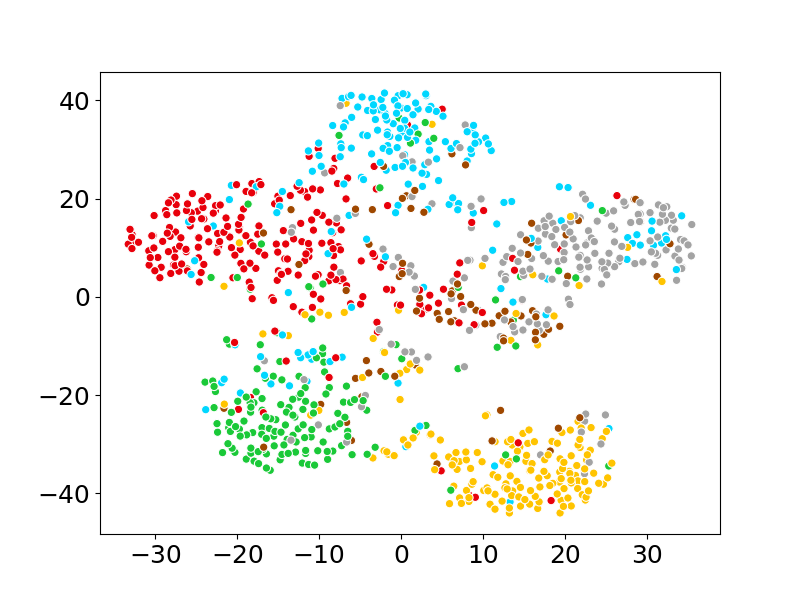}
        \label{fig:visualize_b}
    }
    \subfloat[GraphMix(GCN)]{%
        \includegraphics[width=0.24\linewidth,trim={0cm 0cm 1.5cm  1.5cm},clip]{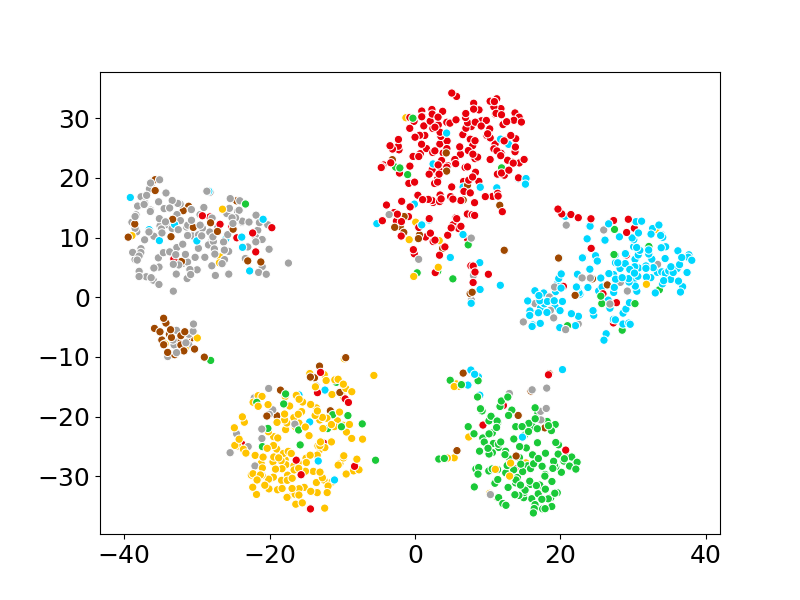}
        \label{fig:visualize_c}
    }
    \subfloat[Class-specific Soft-Rank]{%
       \includegraphics[width=0.24\linewidth,trim={0 0 1cm  1.5cm},clip]{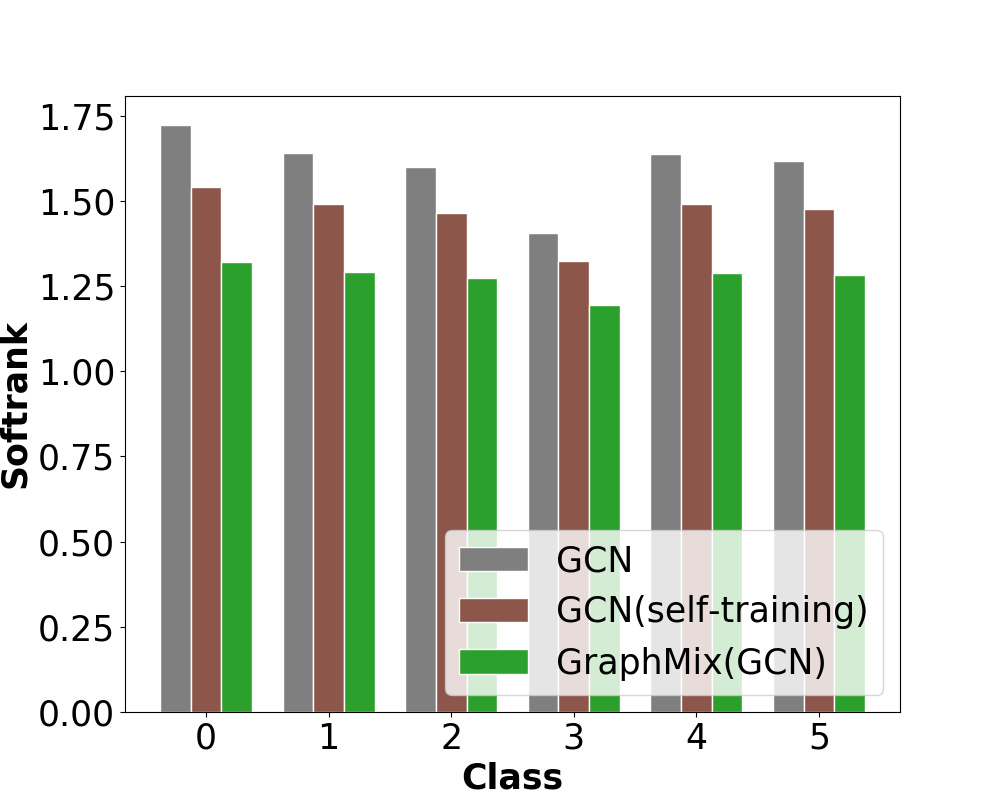}
       \label{fig:visualize_d}
    }
\caption{Two-dimensional representation of the hidden states of Citeseer dataset using (a) GCN, (b) GCN(self-training), (c) GraphMix(GCN), and Soft-Rank (d). GraphMix(GCN) learns better separated representations.  }
\label{fig:visualize}
\end{figure*}


\subsection{Semi-supervised Link Classification}
\label{linkclass}

\begin{table}
\centering
\caption{Results on Link Classification (\%F1 score). $*$ means the results are taken from the corresponding papers.} 
	\label{tab::result_link_classification}
	\resizebox{0.7\columnwidth}{!}{
		\begin{tabular}{c c c}
		        \toprule
		    	\textbf{Algorithm}	& \textbf{Bit OTC} & \textbf{Bit Alpha}  \\ 
		    	
		    \midrule
		    
		     \makecell[c]{DeepWalk \citep{perozzi2014deepwalk}} &63.20 & 62.71 \\
		     
		     \makecell[c]{GMNN* \citep{gmnn}} & \textbf{66.93} &  \textbf{65.86} \\
		     GCN & 65.72$\pm$0.38  & 64.00$\pm$0.19 \\
		     \graphmix(GCN) & 66.35$\pm$0.41 & 65.34$\pm$0.19\\ 
		     \bottomrule
	    \end{tabular}
	}
\end{table}

In the semi-supervised link classification problem, the task is to predict the labels of the remaining links, given a graph and labels of a few links. Following \citep{taskar2004link}, we can formulate the link classification problem as a node classification problem, i.e., given an original graph $G$, we construct a dual Graph $G'$, the node set $V'$ of the dual graph corresponds to the link set $E'$ of the original graph. The nodes in the dual graph $G'$ are connected if their corresponding links in the graph $G$ share a node. The attributes of a node in the dual graph are defined as the index of the nodes of the corresponding link  in the original graph. Using this formulation, we present results on link classification on Bit OTC and Bit Alpha benchmark datasets in the Table \ref{tab::result_link_classification}. As the numbers of  the  positive and negative edges are strongly imbalanced, we report the F1 score. Our results show that \graphmixgcn improves the performance over the baseline GCN method  and is comparable with the recently proposed state-of-the-art method GMNN \citep{gmnn} for both the datasets.


\subsection{Visualization of the Learned Features }

In Figure \ref{fig:visualize}, we present an analysis of the features learned by \graphmix for Cora dataset using the t-SNE \citep{tsne} based 2D visualization of the hidden states. We observe that \graphmix learns hidden states which are better separated and condensed than GCN and GCN(self-training). Here, GCN(self-training) refers to training a GCN in a normal way but with additional self-prediction based targets for the unlabeled samples.  We further evaluate the Soft-rank (refer to Appendix \ref{app:softrank}) of the class-specific hidden states to demonstrate that \graphmixgcn makes the class-specific hidden states more concentrated than GCN and GCN(self-training), as shown in Figure \ref{fig:visualize_d}. Refer to Appendix \ref{app:featureviz} for 2D representation of other datasets.

\section{Related Work}

\begin{itemize}
    \item \textbf{Semi-supervised Learning over Graph Data:} There exists a long line of work for semi-supervised learning over graph data \citep{lp, zhu2003semi}. 
    Contrary to previous methods, the recent GNN based approaches define the convolutional operators using the neighbourhood information of the nodes \citep{kipf2016semi, velivckovic2018graph}.
    These convolution operator based method exhibit state-of-the-results for semi-supervised learning over graph data, hence much of the recent attention is dedicated to proposing architectural changes to these methods \citep{gmnn, u-g-net,ma2019disentangled}. Unlike these methods, we propose a regularization technique that can be applied to any of these GNNs which uses a parameterized transformation on the node features.
    
    \item \textbf{Data Augmentation:} It is well known that the generalization of a learning algorithm can be improved by enlarging the training data size. Since labeling more samples is labour-intensive and costly, Data-augmentation has become \textit{de facto} technique for enlarging the training data size.
    Mixing based algorithms are a particular class of data-augmentation methods in which additional training samples are generated by interpolating the samples (either in the input or hidden space) and/or their corresponding targets. Mixup \citep{mixup}, BC-learning \citep{bclearning}, Manifold Mixup \citep{manifold_mixup}, AMR \citep{AMR} are notable examples of this class of algorithms.
     Unlike, these methods which have been proposed for the fixed topology datasets such as images, in this work, we propose interpolation based data-augmentation techniques for graph-structured data.
     \item \textbf{Regularizing Graph Neural Networks:}
        \label{sec:relatedwork_reg}
        Regularizing Graph Neural Networks has drawn some attention recently. GraphSGAN \citep{graphscan} first uses an embedding method such as DeepWalk \citep{perozzi2014deepwalk} and then trains generator-classifier networks in the adversarial learning setting to generate fake samples in the low-density region between sub-graphs. BVAT \citep{BVAT} and \citep{graphadvtraining} generate adversarial perturbations to the features of the graph nodes while taking graph structure into account. While these methods improve generalization in graph-structured data, they introduce significant additional computation cost: GraphScan requires computing node embedding as a preprocessing step, BVAT and \citep{graphadvtraining} require additional gradient computation for computing adversarial perturbations. Unlike these methods, \graphmix does not introduce any significant additional computation since it is based on  interpolation-based techniques and self-generated targets.

\end{itemize} 

\section{Discussion}
We presented \graphmixns, a simple and efficient regularizer for semi-supervised node classification using graph neural networks. Through extensive experiments, we demonstrated state-of-the-art performances using \graphmix on various benchmark datasets. We also presented a theoritical analysis to compare generalization bounds of \graphmix vs the underlying GNNs. Further, we conduct a systematic ablation study to understand the effect of different components in the performance of \graphmixns.
The strong empirical results of \graphmix suggest that in parallel to designing new architectures, exploring better regularization for graph-structured data is a promising avenue for research. A future research direction is to jointly model the node features and edges of the graph such that it can be further used for generating the synthetic interpolated nodes and their corresponding connectivity to the other nodes in the graph. This will alleviate the need to train the auxiliary FCN in \graphmixns .

\clearpage

\bibliography{icml}
\bibliographystyle{neurips_2020}

\appendix

\section{Appendix}

\subsection{Accurate Target Prediction for Unlabeled data}
\label{sec::predict}
A recently proposed method for accurate target predictions for unlabeled data uses the average of predicted targets across $K$ random augmentations of the input sample \citep{mixmatch}. Along these lines, in GraphMix we compute the predicted-targets as the average of predictions made by GNN on $K$ drop-out versions of the input sample.

Many recent semi-supervised learning algorithms \citep{laine2016temporal, miyato2017vat, meanteacher, ict} are based on the cluster assumption \citep{chapple}, which posits that the class boundary should pass through the low-density regions of the marginal data distribution. One way to enforce this assumption is to explicitly minimize the entropy of the model's predictions $p(y|\mathbf{x}, \theta)$ on unlabeled data by adding an extra loss term to the original loss term \citep{entmin}. The entropy minimization can be also achieved implicitly by modifying the model's prediction on the unlabeled data such that the prediction has low entropy and using these low-entropy predictions as targets for the further training of the model. Examples include "Pseudolabels" \citep{pseudolabel} and "Sharpening" \citep{mixmatch}. Pseudolabeling constructs hard (one-hot) labels for the unlabeled samples which have “high-confidence predictions”. Since many of  the unlabeled samples may have “low-confidence predictions”, they can not be used in the Pseudolabeling technique. On the other hand, Sharpening does not require “high-confidence predictions” , and thus it can be used for all the unlabelled samples. Hence in this work, we use Sharpening for entropy minimization. The Sharpening function over the model prediction $p(y|\mathbf{x}, \theta)$ can be formally defined as follows \citep{mixmatch}, where $T$ is the temperature hyperparameter and $C$ is the number of classes:
\begin{equation}
    \sharpen(p_{i}, T) := p_{i}^{\frac{1}{T}}\bigg/ \sum_{j = 1}^C p_j^{\frac{1}{T}}
    \label{eqn:sharpen}
\end{equation}

\subsection{Connection to Co-training}
\label{sec:cotraining}
The GraphMix approach can be seen as a special instance of the Co-training framework \citep{cotraining}. Co-training assumes that the description of an example can be partitioned into two \textit{distinct} views and either of these views would be sufficient for learning given sufficient labeled data. In this framework, two learning algorithms are trained separately on each view and then the prediction of each learning algorithm on the unlabeled data is used to enlarge the training set of the other. Our method has some important differences and similarities to the Co-training framework. Similar to Co-training, we train two neural networks and the predictions from the GNN are used to enlarge the training set of the FCN. An important difference is that instead of using the predictions from the FCN to enlarge the training set for the GNN, we employ parameter sharing for passing the learned information from the FCN to the GNN. In our experiments, directly using the predictions of the FCN for GNN training resulted in reduced accuracy. This is due to the fact that the number of labeled samples for training the FCN is sufficiently low and hence the FCN does not make accurate enough predictions. Another important difference is that unlike the co-training framework, the FCN and GNN do not use completely distinct views of the data: the FCN uses feature vectors $\mathcal{X}$ and the GNN uses the feature vector and adjacency matrix $ (\mathcal{X}, \mathcal{A})$.  

\subsection{Algorithm}
\label{app:algo}
The procedure for GraphMix training is given in Algorithm \ref{alg:graphmix}.

\begin{algorithm}[t]
        \caption{\graphmix: A procedure for improved training of Graph Neural Networks (GNN)}
   \label{alg:graphmix}
   \begin{algorithmic}[1]
   \footnotesize
   \STATE {\bfseries Input:} A GCN: $G_{\theta}(X,A)$, a FCN: $F_{\theta}(X, \alpha)$ which shares parameters with the GCN. $\betadist$ distribution parameter $\alpha$ for \mm. Number of random perturbations $K$,  Sharpening temperature $T$. maximum value of weight $\gamma$ in the weighted averaging of supervised FCN loss and unsupervised FCN loss. Number of parameter updates $N$. $w(t)$: rampup function for increasing the importance unsupervised loss in FCN training. $(X_L, Y_L)$ represents labeled samples and $X_U$ represents unlabeled samples.

   \FOR{$t = 1$ to $N$}
   \STATE i = random(0,1) \COMMENT{\textit{generate randomly 0 or 1}} \\
   \IF {i=0} 
   \STATE $\mathcal{L}_{\text{supervised}} = \mathcal{L}_{\text{MM}}((\mathbf{X}_l,\mathbf{Y}_l),F_{\theta},\alpha)$ \COMMENT{supervised loss from FCN using the \mm }
   \label{line:MM_sup}\\
   
   \FOR{$k = 1$ to $K$}
   \STATE $\hat{X}_{U, k} = Random Perturbations(X_{U})$ \COMMENT{\textit{Apply $k^{th}$ round of random perturbation to $X_U$}} \label{line:augment_unlabeled} \\
   \ENDFOR
   \STATE $\bar{Y}_{U} = \frac{1}{K}\sum_k g(Y \mid \hat{X}_{U, k}; \theta, A)$ \COMMENT{\textit{Compute average predictions across K perturbations of $X_{U}$} using the GCN} \label{line:average_prediction} \\
   \STATE $Y_{U} = \sharpen(\bar{Y}_{U}, T)$ \COMMENT{\textit{Apply temperature sharpening to the average prediction}} \label{line:sharpen} \\
   \STATE $\mathcal{L}_{\text{unsupervised}} =\mathcal{L}_{\text{MM}}((\mathbf{X}_u,\mathbf{\hat{Y}}_u),F_{\theta},\alpha)$ \COMMENT{unsupervised loss from FCN using the \mm }
   \label{line:MM_unsup}\\
   \STATE $\mathcal{L}= \mathcal{L}_{\text{supervised}} + w(t)*\mathcal{L}_{\text{unsupervised}}$ \COMMENT{\textit{Total loss is the weighted sum of supervised and unsupervised FCN loss}} \label{line:consistency_coeff}\\
   \ELSE
   \STATE  $\mathcal{L} = \mathcal{L} (G_{\theta}(\mathbf{X}_l), \mathbf{Y}_l)$ \COMMENT{\textit{Loss using the vanilla GCN }}
   
   \ENDIF
   \STATE Minimize $\mathcal{L}$ using gradient descent based optimizer such as SGD.
   \ENDFOR
   
\end{algorithmic}
\end{algorithm}

\subsection{Datasets}
\label{app:datasets}
We use three standard benchmark citation network datasets for semi-supervised node classification,  namely Cora, Citeseer and Pubmed. In all these datasets, nodes correspond to documents and edges correspond to citations. Node features correspond to the bag-of-words representation of the document. Each node belongs to one of $C$ classes. During training, the algorithm has access to the feature vectors and edge connectivity of all the nodes but has access to the class labels of only a few of the nodes. 

For semi-supervised link classification, we use two datasets Bitcoin Alpha and Bitcoin OTC from \citep{kumar2016edge, kumar2018rev2}.  The nodes in these datasets correspond to the bitcoin users and the edge weights between them correspond to the degree of trust between the users. Following \citep{gmnn},  we treat edges with weights greater than 3 as positive instances, and edges with weights less than -3 are treated as negative ones. Given a few labeled edges, the task is to predict the labels of the remaining edges. 
The statistics of these datasets as well as the number of training/validation/test nodes is presented in Appendix \ref{app:datasets}.

The statistics of standard benchmark datasets as well as the number of training/validation/test nodes is presented in Table \ref{tab::dataset}. The statistics of larger datasets in given in Table \ref{tab::dataset_big}.

\begin{table}[bht]
	\caption{Dataset statistics.}
	\label{tab::dataset}
	\begin{center}
		{
		\resizebox{1.0\textwidth}{!}{\begin{tabular}{c  c c c c c c c}\hline
		    \textbf{Dataset}  & \textbf{\# Nodes} & \textbf{\# Edges} & \textbf{\# Features} & \textbf{\# Classes} & \textbf{\# Training} & \textbf{\# Validation} & \textbf{\# Test}  \\
	        \midrule
	        Cora &  2,708 & 5,429 & 1,433 & 7 & 140 & 500 & 1,000\\
	        Citeseer  & 3,327 & 4,732 & 3,703 & 6 & 120 & 500 & 1,000 \\
	        Pubmed  & 19,717 & 44,338 & 500 & 3 & 60 & 500 & 1,000\\
	        Bitcoin Alpha  & 3,783 & 24,186 & 3,783 & 2 & 100 & 500 & 3,221 \\
	        Bitcoin OTC  & 5,881 & 35,592 & 5,881 & 2 & 100 & 500 & 5,947 \\
	        \midrule
	    \end{tabular}}
	}
	\end{center}
\end{table}

For larger datasets of Section \ref{app:larger_datasets}, we took processed versions of these dataset available  here \footnote{https://github.com/shchur/gnn-benchmark}. We did 10 random splits of the the data into train/validation/test split. For the classes which had more than 100 samples. We choose 20 samples per class for training, 30 samples per class for validation and the remaining samples as test data. For the classes which had less than 100 samples, we chose 20\% samples, per class for training, 30\% samples for validation and the remaining for testing. For each split we run experiments using 100 random seeds. The statistics of these datasets is presented in Appendix Table \ref{table:dataset_new_statistics} and 

\begin{table}[h!]
    \caption{Dataset statistics for Larger datasets
    }
    \label{tab::dataset_big}
    \begin{center}
        \begin{tabular}{l r r r r }
            \toprule
            \textbf{Datasets}    & \textbf{Classes} & \textbf{Features} & \textbf{Nodes} & \textbf{Edges}  \\
            \midrule
            
            Cora-Full     &           67 &          8710 &      18703 &      62421 \\
            Coauthor-CS       &           15 &          6805 &      18333 &      81894  \\
            Coauthor-Physics        &            5 &          8415 &      34493 &     247962  \\
            NELL                &        210    &            5414        & 65755     & 266144 \\
            \bottomrule
           
        \end{tabular}
    \end{center}
    
    \label{table:dataset_new_statistics}
\end{table}

\subsection{Ablation Study}
\label{subsection:ablation}
Since \graphmix consists of various components, some of which are common with the existing literature of semi-supervised learning, we set out to study the effect of various components by systematically removing or adding a component from \graphmix. We measure the effect of the following:

\begin{itemize}
    \item Removing both the Manifold Mixup and predicted targets from the FCN training.
    \item Having only the Manifold Mixup training for FCN ( No predicted targets for FCN training)
    \item Using the predicted targets for the FCN training (No Manifold Mixup in FCN training)
    \item Using both Manifold Mixup and predicted targets for FCN training
\end{itemize}

\begin{table*}[]
\caption{Ablation study results using 10 labeled samples per class (\% test accuracy). We report mean and standard deviation over ten trials. See Section \ref{subsection:ablation} for the meaning of methods in leftmost column.}
\centering
\begin{tabular}{lrrr}
\toprule
\textbf{Ablation}	& \textbf{Cora} & \textbf{Citeseer} & \textbf{Pubmed} \\ 
\midrule

\makecell[l]{ Without Manifold Mixup and\\ without predicted-targets} & 68.78$\pm$3.54  & 61.01$\pm$1.24 & 72.56$\pm$1.08  \\
With Predicted targets & 69.08$\pm$5.03 & 62.66$\pm$1.80 & 73.07$\pm$0.94  \\
With Manifold Mixup & 73.55$\pm$3.29  & 65.72$\pm$2.10  &  75.74$\pm$1.69 \\
With Manifold Mixup and Predicted targets & 79.30$\pm$1.36  & 70.78$\pm$1.41  & 77.13$\pm$3.60   \\



\bottomrule
\end{tabular}
\vskip 0.1in
\label{tab:ablation}
\vskip -0.2in
\end{table*}

The ablation results for semi-supervised node classification are presented in Table \ref{tab:ablation}. We did not do any hyperparameter tuning for the ablation study and used the best performing hyperparameters found for the results presented in Table \ref{tab::standard_split}. We observe that all the components of \graphmix contribute to its performance, with Manifold Mixup training of FCN contributing possibly the most. 

\subsection{Comparison with State-of-the-art Methods}
\label{app:standard_split}
We present the comparion of \graphmix with the recent state-of-the-art methods as well as earlier methods using the standard Train/Validation/Test split in Table \ref{tab:full}. We additionally use self-training based baselines, where FCN, GCN, GAT and Graph-U-Net are trained with self-generated targets. These are named as FCN (self-training), GCN (self-training), GAT (self-training) and Graph-U-Net(self-training) respectively in Table ~\ref{tab:full}. For generating the predicted-targets in above two baselines, we followed the procedure of Appendix ~\ref{sec::predict}.

\begin{table*}[ht!]
\caption{Comparison of GraphMix with other methods (\% test accuracy ), for Cora, Citeseer and Pubmed.}
\label{tab:full}
\begin{center}
\begin{tabular}{l l l l}
\toprule
{\bf Method} & {\bf Cora} & {\bf Citeseer} & {\bf Pubmed}\\ \midrule
\midrule
Results reported from the literature & & & \\
\midrule
\midrule
MLP & 55.1\% & 46.5\% & 71.4\% \\
ManiReg \citep{belkin2006manifold} & 59.5\% & 60.1\% & 70.7\%\\
SemiEmb \citep{weston2012deep} & 59.0\% & 59.6\% & 71.7\%\\
LP \citep{zhu2003semi} & 68.0\% & 45.3\% & 63.0\%\\
DeepWalk \citep{perozzi2014deepwalk} & 67.2\% & 43.2\% & 65.3\%\\
ICA \citep{lu2003link} & 75.1\% & 69.1\% & 73.9\%\\
Planetoid \citep{yang2016revisiting} & 75.7\% & 64.7\% & 77.2\%\\
Chebyshev \citep{defferrard} & 81.2\% & 69.8\% & 74.4\%\\
GCN \citep{kipf2016semi} & 81.5\% & 70.3\% &  79.0\%\\
MoNet \citep{monti2016geometric} & 81.7 $\pm$ 0.5\% & --- & 78.8 $\pm$ 0.3\%\\ 
 GAT \citep{velivckovic2018graph}&  83.0 $\pm$ 0.7\% &  72.5 $\pm$ 0.7\% &  79.0 $\pm$ 0.3\%\\
GraphScan \citep{graphscan} & 83.3 $\pm$1.3 & 73.1$\pm$1.8 & --- \\
DisenGCN  \citep{ma2019disentangled} & 83.7\% & 73.4\% & 80.5\% \\
Graph U-Net  \citep{u-g-net} & \textbf{84.4\%} & 73.2\% & 79.6\% \\
BVAT \citep{BVAT} & 83.6$\pm$0.5 & \textbf{74.0$\pm$0.6} & 79.9$\pm$0.4 \\

\midrule
\midrule
Our Experiments & & & \\
\midrule
\midrule

 GCN & 81.30$\pm$0.66  & 70.61$\pm$0.22 & 79.86$\pm$0.34\\
 GAT & 82.70$\pm$0.21  & 70.40$\pm$0.35 & 79.05$\pm$0.64 \\
 
 Graph U-Net & 81.74$\pm$0.54  & 67.69$\pm$1.10 & 77.73 $\pm$0.98 \\
 
 FCN (self-training) &  80.30$\pm$0.75 & 71.50$\pm$0.80 & 77.40$\pm$0.37\\
 GCN (self-training) &  82.03$\pm$0.43 & 73.38$\pm$0.35 & 80.42$\pm$0.36\\
 GAT (self-training) & 82.95$\pm$0.23 & 72.87$\pm $0.51 & 79.67$\pm$0.69 \\
 Graph-U-Net (self-training) & 82.01$\pm$0.91   & 68.23$\pm$1.57 & 78.12$\pm$1.23 \\

 \midrule 
 
 \graphmix(GCN) & \textbf{83.94$\pm$0.57}  & \textbf{74.52$\pm$0.59} & 80.98$\pm$0.55 \\ 
 \graphmix(GAT) & 83.32$\pm$0.18  & 73.08$\pm$0.23 & \textbf{81.10$\pm$0.78} \\
 
 \graphmix(Graph U-Net) & 82.18$\pm$0.63  & 69.00 $\pm$1.32 & 78.76$\pm$1.09 \\
 
 \bottomrule
\end{tabular}
\end{center}
\end{table*}

\subsection{Results with fewer labeled samples}
\label{sec:fewerlabels}
We further evaluate the effectiveness of \graphmix in the learning regimes where fewer labeled samples exist. For each class, we randomly sampled $K\in \{5, 10\}$ samples for training and the same number of samples for the validation. We used all the remaining labeled samples as the test set. We repeated this process for $10$ times. The results in Table \ref{tab::results_less_labeled} show that \graphmix achieves even better improvements when the labeled samples are fewer.  

\begin{table*}[bht]
        \caption{Results using less labeled samples (\% test accuracy). $K$ referes to the number of labeled samples per class.}
	\label{tab::results_less_labeled}
	\begin{center}
	{
		\resizebox{\textwidth}{!}{\begin{tabular}{c c c c c c c}\hline
		    	\multirow{2}{*}{\textbf{Algorithm}}	& \multicolumn{2}{c}{\textbf{Cora}} & \multicolumn{2}{c}{\textbf{Citeseer}} & \multicolumn{2}{c}{\textbf{Pubmed}} \\ 
		     & $K=5$& $K=10$ & $K=5$ &$K=10$ &$K=5$ & $K=10$ \\
		    \midrule

		     GCN & 66.39$\pm$4.26  & 72.91$\pm$3.10 & 55.61$\pm$5.75 & 64.19$\pm$3.89 &66.06$\pm3.85$ & 75.57$\pm$1.58 \\
		     
		     GAT & 68.17$\pm$5.54 & 73.88$\pm$4.35 &55.54$\pm$1.82 & 61.63$\pm$0.42 &  64.24$\pm$4.79 & 73.60$\pm$1.85 \\
		     
		     Graph U-Net &64.42$\pm$5.44 & 71.48$\pm$3.03 & 49.43$\pm$5.81 & 61.16$\pm$3.47 &65.05$\pm$4.69  &  68.65$\pm$3.69 \\
		     
		     \midrule
		     \graphmix(GCN) & \textbf{71.99$\pm$6.46}  & \textbf{79.30$\pm$1.36} & \textbf{58.55$\pm$2.26} & \textbf{70.78$\pm$1.41}& \textbf{67.66$\pm$3.90}  & \textbf{77.13$\pm$3.60} \\
		     
		     \graphmix(GAT) &72.01$\pm$6.68  & 75.82$\pm$2.73 & 57.6$\pm$0.64 & 62.24$\pm$2.90  & 66.61$\pm$3.69 & 75.96$\pm$1.70 \\
		     \graphmix(Graph U-Net) &66.84$\pm$6 5.10 &73.14$\pm$3.17 & 54.39$\pm$5.07 & 64.36$\pm$3.48   & 67.40$\pm$5.33 &70.43$\pm$3.75  \\
		     
		     \midrule
	    \end{tabular}}
	}
	\end{center}
\end{table*}

\subsection{Implementation and Hyperparameter Details}
\label{app:hyper}
We use the standard benchmark architecture as used in GCN \citep{kipf2016semi}, GAT \citep{velivckovic2018graph} and GMNN \citep{gmnn}, among others. This architecture has one hidden layer and the graph convolution is applied twice : on the input layer and on the output of the hidden layer. The FCN in \graphmix shares the parameters with the GCN.

\graphmix introduces four additional hyperparameters, namely the $\alpha$ parameter of $\betadist$ distribution used in Manifold Mixup training of the FCN, the maximum weighing coefficient $\mathcal{\gamma}$ which controls the trade-off between the supervised loss and the unsupervised loss (loss computed using the predicted-targets) of FCN, the temparature $T$ in sharpening and the number of random perturbations $K$ applied to the input data for the averaging of the predictions.

We conducted minimal hyperparameter seach over only $\alpha$ and  $\mathcal{\gamma}$ and fixed the hyperparameters $T $ and $K$ to $0.1$ and $10$ respectively. The other hyperparameters were set  to the best values for underlying GNN (GCN or GAT), including the learning rate, the $L2$ decay rate, number of units in the hidden layer etc. We observed that \graphmix is not very sensitive to the values of $\alpha$ and $\mathcal{\gamma}$ and similar values of these hyperparameters work well across all the benchmark datasets. Refer to Appendix \ref{app:hyper} and \ref{app:hyper_selection}  for the details about the hyperparameter values and the procedure used for the best hyperparameters selection.

\subsubsection{For results reported in Section \ref{subsec:results}}
\label{sec:hyper_main}
 
For GCN and GraphMix(GCN), we used Adam optimizer with learning rate $0.01$ and $L2$-decay 5e-4, the number of units in the hidden layer $16$ , dropout rate in the input layer and hidden layer was set to $0.5$ and $0.0$, respectively. For GAT and GraphMix(GAT), we used Adam optimizer with learning rate $0.005$ and $L2$-decay 5e-4, the number of units in the hidden layer $8$ , and the dropout rate in the input layer and hidden layer was searched from the values  $\{0.2, 0.5, 0.8\}$.

For $\alpha$  and $\mathcal{\gamma}$ of \graphmixgcn and \graphmixgat, we searched over the values in the set $[0.0, 0.1, 1.0, 2.0]$ and  $[0.1, 1.0, 10.0, 20.0]$  respectively.

For \graphmixgcn: $\alpha=1.0$ works best across all the datasets. 
$\mathcal{\gamma}=1.0$ works best for Cora and Citeseer and $\mathcal{\gamma}=10.0$ works best for Pubmed.

For \graphmixgat: $\alpha=1.0$ works best for Cora and Citeseer and  $\alpha=0.1$ works best for Pubmed. $\mathcal{\gamma}=1.0$ works best for Cora and Citeseer and $\mathcal{\gamma}=10.0$ works best for Pubmed. Input droputrate=0.5  and hidden dropout rate=0.5 work best for Cora and Citeseer and Input dropout rate=0.2 and hidden dropout rate =0.2 work best for Pubmed.

We conducted all the experiments for 2000 epochs. The value of weighing coefficient $w(t)$ in Algorithm \ref{alg:graphmix}) is increased from $0$ to its maximum value $\gamma$ from epoch 500 to 1000 using the sigmoid ramp-up of Mean-Teacher \citep{meanteacher}. 

\subsubsection{Hyperparameter Details for Results in Section \ref{app:larger_datasets}}
\label{app:hyper_largerdata}
For all the experiments we use the standard architecture mentioned in Section \ref{app:hyper} and used Adam optimizer with learning rate 0.001 and 64 hidden units in the hidden layer. For Coauthor-CS and Coauthor-Physics, we trained the network for 2000 epochs. For Cora-Full, we trained the network for 5000 epochs because we observed the training loss of Cora-Full dataset takes longer to converge.

For Coauthor-CS and Coauthor-Physics:  We set the input layer dropout rate to 0.5 and weight-decay to 0.0005, both for GCN and \graphmixgcn. We did not conduct any hyperparameter search over the \graphmix hyperparameters $\alpha$, $\lambda_{max}$, temparature $T$ and number of random permutations $K$ applied to the input data for \graphmixgcn for these two datasets, and set these values to $1.0$, $1.0$, $0.1$ and $10$ respectively.  

For Cora-Full dataset: We found  input layer dropout rate  0.2 and weight-decay  0.0 to be the best for both GCN and \graphmixgcn.
For \graphmixgcn we fixed $\alpha$,  temparature $T$ and number of random permutations $K$ to $1.0$ $0.1$ and $10$ respectively. For  $\lambda_{max}$, we did search over $\{1.0, 10.0, 20.0\}$ and found that $10.0$ works best.

For all the \graphmixgcn experiments, the value of weighing coefficient $w(t)$  in Algorithm \ref{alg:graphmix}) is increased from $0$ to its maximum value $\gamma_{max}$ from epoch 500 to 1000 using the sigmoid ramp-up of Mean-Teacher \citep{meanteacher}. 

\subsubsection{For results reported in Section \ref{linkclass}}
For $\alpha$ of \graphmixgcn, we searched over the values in the set $[0.0, 0.1, 0.5, 1.0]$ and found that $0.1$ works best for both the datasets. For $\mathcal{\gamma}$, we searched over the values in the set $[0.1, 1.0, 10.0]$ and found that $0.1$ works best for both the datasets. We conducted all the experiments for 150 epochs. The value of weighting coefficient $w(t)$  in Algorithm \ref{alg:graphmix}) is increased from $0$ to its maximum value $\mathcal{\gamma}$ from epoch 75 to 125 using the sigmoid ramp-up of Mean-Teacher \citep{meanteacher}.

Both for \graphmixgcn and GCN, we use Adam optimizer with learning rate $0.01$ and $L2$-decay $0.0$, the number of units in the hidden layer $128$ , dropout rate in the input layer  was set to $0.5$.

\subsubsection{For results reported in Section \ref{sec:fewerlabels}}
For $\alpha$ of \graphmixgcn, we searched over the values in the set $[0.0, 0.1, 0.5, 1.0]$ and found that $0.1$ works best across all the datasets. For $\mathcal{\gamma}$, we searched over the values in the set $[0.1, 1.0, 10.0]$ and found that $0.1$ and $1.0$ works best across all the datasets. Rest of the details for \graphmixgcn and GCN are same as Section \ref{sec:hyper_main}.

\subsection{Hyperparameter Selection}
\label{app:hyper_selection}

For each configuration of hyperparameters, we run the experiments with $100$ random seeds. We select the hyperparameter configuration which has the best validation accuracy averaged over these $100$ trials. With this best hyperparameter configuration, for $100$ random seeds,  we train the model again and use the validataion set for model selection ( i.e. we report the test accuracy at the epoch which has best validation accuracy.)

\begin{figure*}
    \centering
    \includegraphics[width=0.32\linewidth,trim={1.1cm 0.5cm 1.6cm 1.4cm},clip]{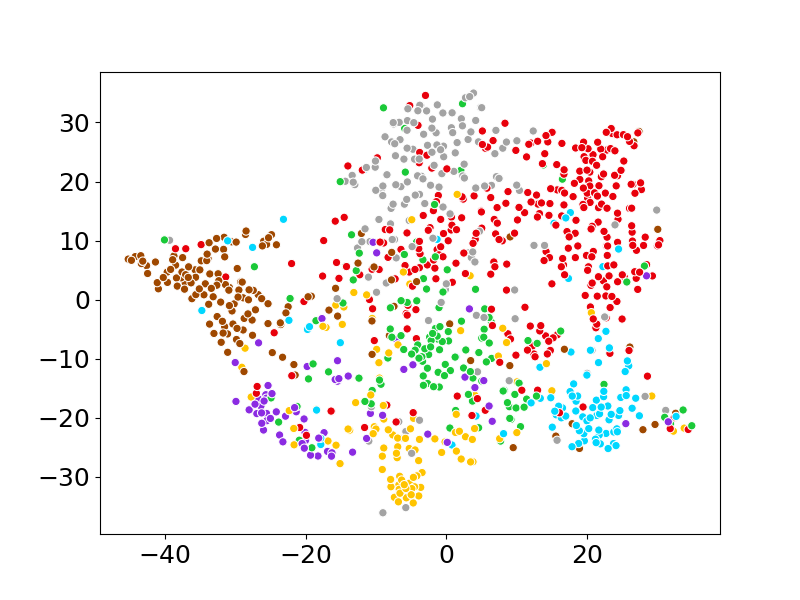}
    \includegraphics[width=0.32\linewidth,trim={1.1cm 0.5cm 1.6cm 1.4cm},clip]{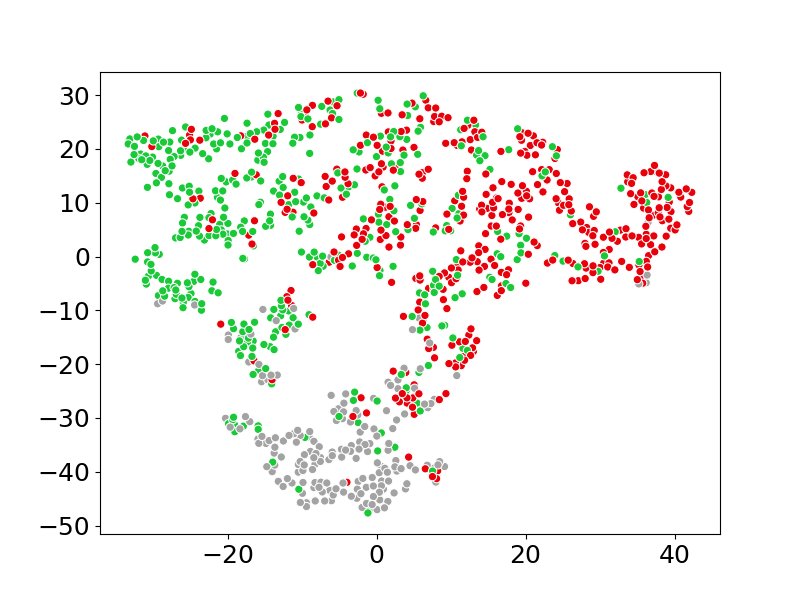}
    \includegraphics[width=0.32\linewidth,trim={1.1cm 0.5cm 1.6cm 1.4cm},clip]{figures/featureviz/citeseer_gcn.png}
    \includegraphics[width=0.32\linewidth,trim={1.1cm 0.5cm 1.6cm 1.4cm},clip]{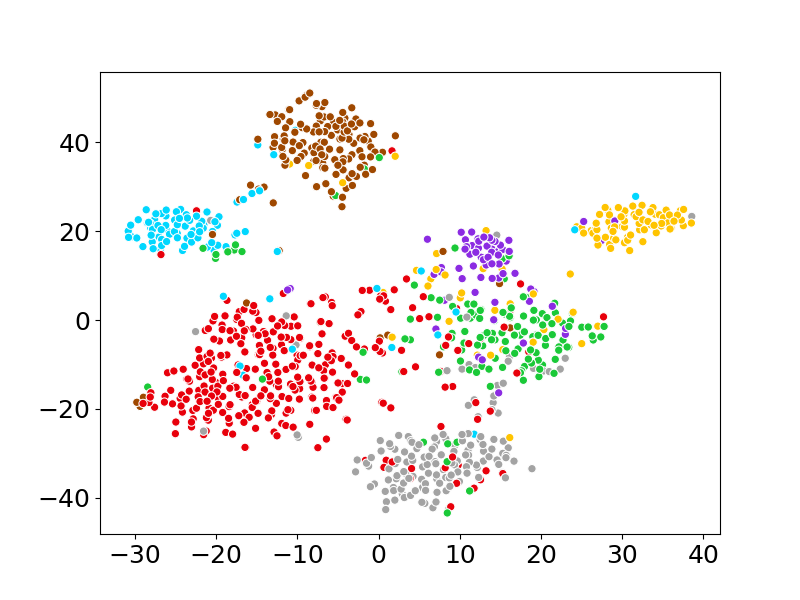}
    \includegraphics[width=0.32\linewidth,trim={1.1cm 0.5cm 1.6cm 1.4cm},clip]{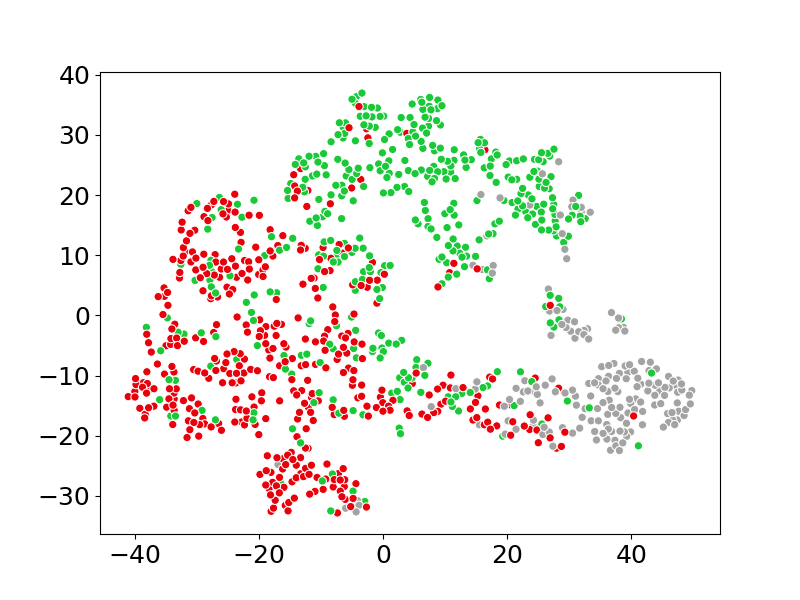}
    \includegraphics[width=0.32\linewidth,trim={1.1cm 0.5cm 1.6cm 1.4cm},clip]{figures/featureviz/citeseer_graphmix.png}
    \caption{T-SNE of hidden states for Cora (left), Pubmed (middle), and Citeseer (right).  Top row is GCN baseline, bottom row is GraphMix.}
    \label{fig:fullfeatureviz}
\end{figure*}

\begin{figure*}
    \centering
    \includegraphics[width=0.45\linewidth]{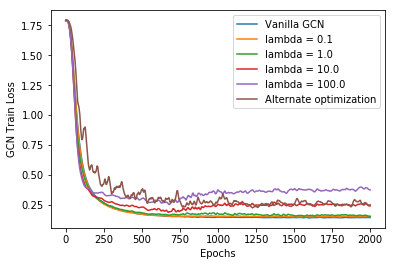}
    \includegraphics[width=0.45\linewidth]{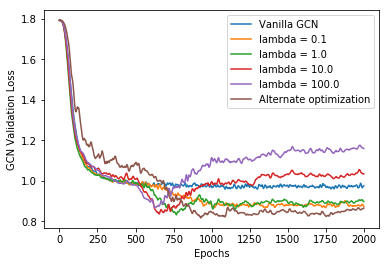}
    \caption{GCN train loss and validation loss for Alternate optimization vs. weighted joint optimization. lambda = X.X represents the value of $\lambda$ for simultaneous optimization in Eq \ref{eq:graphmix_loss}. }
    \label{fig:alternate_optimization}
\end{figure*}

\subsection{Soft-Rank}
\label{app:softrank}
Let $\textbf{H}$ be a matrix containing the hidden states of all the samples from a particular class. The Soft-Rank of  matrix $\textbf{H}$ is defined by the sum of the singular values of the matrix divided by the largest singular value.  A lower Soft-Rank implies fewer dimensions with substantial variability and it provides a continuous analogue to the notion of rank from matrix algebra.  This provides evidence that the concentration of class-specific states observed when using GraphMix in Figure~\ref{fig:fullfeatureviz} can be measured directly from the hidden states and is not an artifact of the T-SNE visualization.

\subsection{Feature Visualization}
\label{app:featureviz}
We present the 2D visualization of the hidden states learned using GCN and GraphMix(GCN) for Cora, Pubmed and Citeseer datasets in Figure \ref{fig:fullfeatureviz}. We observe that for Cora and Citeseer, GraphMix learns substantially better hidden states than GCN. For Pubmed, we observe that although there is no clear separation between classes, "Green" and "Red" classes overlap less using the GraphMix, resulting in better hidden states.

\subsection{Joint Optimization vs Alternate optimization}
\label{app:joint_alternate}

In this Section, we discuss the effect of hyperparameter $\lambda$, that is used to compute the weighted sum of GCN and FCN losses in in Eq \ref{eq:graphmix_loss}. In Figure \ref{fig:alternate_optimization}, we see that a wide range of $\lambda$ ( from 0.1 to 1.0) achieves better validation accuracy than the vanilla GCN. Furthermore, alternate optimization of the FCN loss and the GCN loss achieves substantially better validation accuracy than the simultaneous optimization.

\end{document}